\documentclass[numbers]{cas-dc}

\usepackage{natbib}
\usepackage{booktabs}
\usepackage{amsmath,amssymb}
\usepackage{graphicx}
\usepackage{threeparttable}
\graphicspath{{figs/}}

\usepackage{siunitx}
\usepackage{newtxtext}

\usepackage[font=small,labelfont=bf,skip=4pt]{caption}

\usepackage{xcolor}
\usepackage{hyperref}

\hypersetup{
    colorlinks = true,
    citecolor  = blue,
    linkcolor  = black,
    urlcolor   = blue
}

\begin{document}

\title{DSR-Net: Learning Dynamic Skeletal Relations for 3D Human Pose Estimation}

\author[1]{Fan Li}
\author[1]{Jian Xiang}\corref{cor1} 
\author[1]{Linlin Xue}
\author[1]{Yuwei Gao}

\cortext[cor1]{Corresponding author. Email: xiangjian@zust.edu.cn}

\address[1]{School of Computer Science and Technology, Zhejiang University of Science and Technology, Liuxia, Hangzhou 310023, China}

\begin{abstract}
3D human pose estimation requires accurate modeling of motion-dependent skeletal structures and inter-joint dependencies. Existing methods typically rely on fixed skeletal topologies in Graph Convolutional Networks (GCNs) or implicitly capture joint relations through attention mechanisms, limiting their adaptability to diverse motion patterns.

To address this issue, we propose DSR-Net, a dynamic structural representation learning framework for 3D human pose estimation. Specifically, a Dynamic Skeletal Relation (DSR) module is introduced to learn adaptive inter-joint dependencies beyond predefined skeletal connections. Based on the learned relations, a hierarchical relation-aware fusion strategy is developed to facilitate effective interaction between structural and temporal representations. In addition, a lightweight Residual Skeletal Recalibration Module (RSRM) is designed to enhance the stability of structure-aware features with minimal computational overhead.

Extensive experiments on Human3.6M and MPI-INF-3DHP demonstrate that DSR-Net achieves MPJPEs of 37.9 mm and 14.9 mm, respectively, consistently improving over strong baselines. The results validate the effectiveness of dynamic skeletal relation modeling for robust 3D human pose estimation.
\end{abstract}


\maketitle

\section{Introduction}

The goal of 3D human pose estimation is to recover the spatial positions and motion trajectories of human joints from temporal observations. Beyond accurate joint coordinate regression, the task also requires effective modeling of motion-dependent inter-joint relations within the skeletal structure. Since these relations vary with pose configurations and dynamic motion states, learning adaptive skeletal dependencies is critical for accurate 3D pose estimation.

Recent advances in deep representation learning have significantly improved the capability of learning complex spatiotemporal representations for human motion analysis and 3D pose estimation~\cite{lin_unsupervised_2024}. However, most existing methods still represent skeletal dependencies using fixed topologies or implicitly learned relations, limiting their ability to capture motion-dependent structural variations.

Recently, Transformer-based architectures~\cite{vaswani_attention_2017} have been widely adopted for 3D human pose estimation due to their strong global modeling capability. By leveraging self-attention mechanisms, Transformers can effectively capture long-range temporal dependencies and cross-joint relations. However, the learned dependencies are often implicitly encoded within attention weights, making it difficult to explicitly characterize motion-dependent structural relations and local skeletal variations. In contrast, graph convolutional network (GCN)-based methods explicitly model skeletal topology and are capable of capturing local inter-joint relations, offering advantages in incorporating structural priors. Nevertheless, conventional GCNs typically rely on fixed or weakly adaptive adjacency matrices, making these methods insufficient for learning dynamic skeletal relations that evolve with motion.

To combine the global modeling capability of Transformers and the structural modeling advantages of GCNs, recent studies have explored hybrid Transformer–GCN architectures that integrate temporal modeling with spatial structure modeling. Although existing hybrid Transformer-GCN methods have achieved improved performance, most still rely on static or weakly adaptive structural priors, limiting their ability to model motion-dependent skeletal relations and adaptive cross-stream interactions under diverse motion patterns. Consequently, dynamically evolving skeletal structures are not explicitly represented, which limits the ability of existing methods to adapt to diverse motion patterns. As a consequence, feature interaction across different modeling branches may become insufficiently adaptive, while additional structural modeling modules may also introduce redundant representations and affect training stability.

In this paper, we propose DSR-Net, a hybrid Graph-Transformer framework for 3D human pose estimation centered on Dynamic Skeletal Relation (DSR) modeling. The core idea of DSR-Net is to dynamically learn motion-dependent joint dependencies beyond fixed skeletal topology, enabling adaptive structural representations that explicitly characterize skeletal relation variations under different motion states.

To effectively exploit the dynamically learned skeletal relations, we further design a hierarchical relation-aware fusion strategy with adaptive gated interaction and Dynamic Channel Fusion (DCF) for enhanced cross-stream feature interaction and spatiotemporal representation learning. In addition, a lightweight Residual Skeletal Recalibration Module (RSRM) is introduced to stabilize and refine dynamically learned structural representations while maintaining computational efficiency.

Unlike existing methods that rely on predefined skeletal topology or implicit attention patterns, DSR-Net explicitly models dynamic skeletal relations as adaptive structural representations for motion-aware pose estimation. The main contributions of this paper are summarized as follows:
\begin{itemize}

\item[(1)] We propose a Dynamic Skeletal Relations (DSR) modeling framework that dynamically learns motion-dependent joint dependencies beyond fixed skeletal topology, enabling adaptive structural representation learning for 3D human pose estimation.

\item[(2)] To effectively exploit dynamically learned skeletal relations, we design a hierarchical relation-aware fusion strategy with adaptive gated interaction and Dynamic Channel Fusion (DCF) to enhance cross-stream feature interaction and spatiotemporal representation learning.

\item[(3)] We introduce a lightweight Residual Skeletal Recalibration Module (RSRM) to stabilize and refine dynamically learned structural representations with limited computational overhead.

\end{itemize}

\section{Related Work}

\subsection{2D-to-3D Pose Lifting}

Monocular 2D-to-3D pose lifting has become a dominant
research direction in 3D human pose estimation due to its low deployment cost and broad applicability. Unlike direct regression methods from RGB images, 2D-to-3D lifting methods first estimate 2D joint coordinates and then map them into 3D space, effectively decoupling visual perception from
structural representation learning. Martinez et al.
\cite{martinez_simple_2017} demonstrated the effectiveness of this paradigm, which has since become a widely adopted baseline framework.

Early lifting methods mainly relied on temporal convolution networks to model short-term motion continuity \cite{pavllo_3d_2019}. More recently, Transformer-based methods have shown strong capability in capturing long-range temporal dependencies through self-attention mechanisms,
while Graph Convolutional Networks (GCNs) explicitly model skeletal topology and inter-joint structural relations.
Based on these advantages, recent studies increasingly adopt hybrid Transformer-GCN frameworks for joint spatiotemporal representation learning.

\subsection{Spatiotemporal Structural Modeling}

Recent 2D-to-3D pose lifting methods mainly improve pose representation learning through Transformer-based temporal modeling and GCN-based structural modeling strategies. Transformer architectures exhibit strong capability in capturing long-range temporal dependencies through self-attention mechanisms, while GCNs explicitly model skeletal topology and inter-joint structural relations.

Transformer-based methods such as MHFormer
\cite{li_mhformer_2021}, MixSTE
\cite{zhang_mixste_2022}, MotionBERT
\cite{zhu_motionbert_2023}, DSTFormer
\cite{zhang_dstformer_2024}, PoseFormerV2
\cite{zhao_poseformerv2_2023}, and STCFormer
\cite{tang_3d_2023} improve long-range temporal
dependency modeling through attention mechanisms,
frequency representation learning, or structured
temporal modeling strategies.

Recent studies further explore state-space modeling,
adaptive pose encoding, and motion priors to enhance
temporal representation capability
\cite{huang_posemamba_2025,
peng_ktpformer_2024,
zhang_stapformer_2024}.

GCN-based methods explicitly model skeletal structures using graph representations of human joints and bones
\cite{kipf_semi-supervised_2016}.
Representative approaches improve pose estimation through semantic graph learning, adaptive topology modeling, global-local structural interaction, and graph optimization strategies
\cite{cai_exploiting_2019,
zhao_semantic_2019,
xu_graph_2021,
liu2020Disentangling,
yu_gla-gcn_2023,
zou_modulated_2021,
lin_origin_2025,
ci_optimizing_2019}.

Although Transformer-based methods achieve strong temporal modeling performance and GCN-based methods provide explicit structural constraints, existing approaches still exhibit limitations in jointly modeling motion-dependent skeletal relations and adaptive spatiotemporal interactions.
Transformer architectures often model joint relations
implicitly, while many GCN-based methods rely on static or globally shared graph topologies, limiting their flexibility under complex motion scenarios.

These limitations have motivated the development of hybrid Transformer-GCN frameworks for more comprehensive spatiotemporal representation learning in 3D human pose estimation.

\subsection{Hybrid Spatiotemporal Modeling}

Recent hybrid Transformer-GCN frameworks attempt to jointly model temporal dynamics and skeletal structures for 3D human pose estimation. Existing studies mainly improve hybrid representation learning through architectural integration and cross-branch feature fusion strategies.

Some methods explicitly incorporate graph structures into Transformer architectures to enhance structural modeling capability \cite{liu_stgformer_2026}.
Other approaches perform adaptive feature fusion between Transformer and GCN branches to improve spatiotemporal representation learning
\cite{mehraban_motionagformer_2024,
lu_cgfusionformer_2025,
zheng_ehgformer_2025,
duan_dual-stream_2026}.
In addition, diffusion-based methods have also demonstrated promising performance in robust 3D pose estimation
\cite{gong_diffpose_2023}.

Despite these advances, most existing hybrid frameworks still rely on static skeletal priors or relatively fixed fusion strategies, limiting their ability to capture motion-dependent joint relations and adaptive cross-stream interactions under complex motion scenarios.

To address these limitations, this work proposes a dynamic skeletal relation learning framework with adaptive feature fusion for more flexible spatiotemporal representation learning in 3D human pose estimation.

\section{Method}

\begin{figure*}[htbp]
  \centering
  \includegraphics[width=0.9\textwidth]{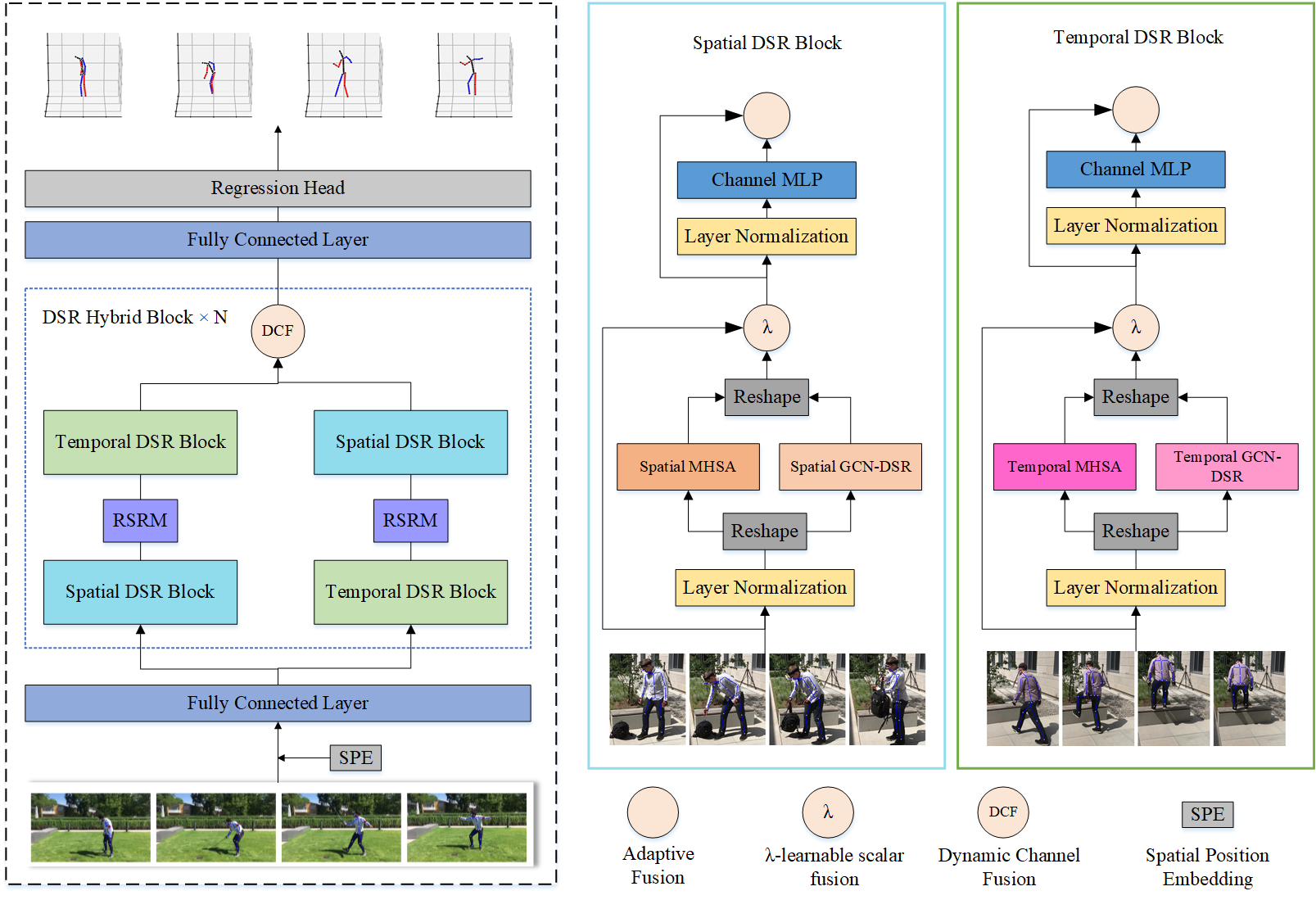}
  \caption{Overall architecture of the proposed framework.}
  \label{fig:module}
\end{figure*}

\subsection{Overall Architecture}

The proposed DSR-Net is a dynamic skeletal relation learning framework for 3D human pose estimation. Built upon the dual-stream architecture of MixTGFormer~\cite{duan_dual-stream_2026}, DSR-Net introduces Dynamic Skeletal Relations (DSR) modeling, Residual Skeletal Recalibration (RSRM), and hierarchical relation-aware fusion to enable adaptive structural representation learning under diverse motion patterns.

Unlike conventional methods that rely on fixed skeletal topology, the proposed framework models inter-joint dependencies in a data-driven manner, enabling dynamic graph construction under varying motion patterns. By capturing motion-dependent skeletal relations, DSR-Net effectively models both spatial configurations and temporal dynamics within a unified Transformer-GCN framework.

As illustrated in Fig.~\ref{fig:module}, DSR-Net maps a sequence of 2D keypoints to the corresponding 3D pose sequence through a dual-stream DSR architecture, consisting of a \textit{Spatial-First} stream and a \textit{Temporal-First} stream. These two streams capture complementary spatiotemporal dependencies from different modeling perspectives. Each stage is composed of stacked DSR Hybrid Blocks, where spatial and temporal representations are progressively refined through adaptive relation modeling and cross-stream interaction.

To instantiate the proposed Dynamic Skeletal Relations (DSR) modeling framework, a graph-based dynamic relation reasoning module (GCN-DSR) is introduced to explicitly construct motion-dependent skeletal relations as adaptive structural representations. Specifically, GCN-DSR dynamically learns motion-aware inter-joint dependencies beyond predefined skeletal topology, enabling adaptive structural modeling under diverse motion patterns.

Meanwhile, a Residual Skeletal Recalibration Module (RSRM) is incorporated to stabilize dynamically learned structural representations through lightweight structure-guided feature recalibration.

Furthermore, a hierarchical relation-aware fusion strategy is designed to progressively integrate complementary spatiotemporal information. Specifically, an intra-block gated fusion mechanism is employed to adaptively combine attention-based global representations with graph-based local structural features, while a Dynamic Channel Fusion (DCF) mechanism is employed within each DSR Hybrid Block for dynamic cross-stream interaction. Through the hierarchical coordination of local structural fusion and cross-stream feature interaction, the proposed framework achieves more robust and consistent spatiotemporal representations. Finally, the regression head outputs the estimated 3D joint coordinates.

\textbf{Hybrid spatiotemporal branches.} 
The model consists of two parallel computational paths: one follows a “spatial-to-temporal” modeling order, while the other follows a “temporal-to-spatial” order. This dual-stream design enables the model to capture complementary spatiotemporal dependencies under different modeling sequences.

In the spatial modeling stage, input features are processed in parallel by Spatial Multi-Head Self-Attention (S-MHSA) and the DSR-guided graph reasoning branch (GCN-DSR). S-MHSA captures long-range global dependencies among joints within a single frame, while GCN-DSR dynamically learns adaptive inter-joint relations based on motion-dependent feature interactions. This enables flexible structural modeling beyond predefined skeletal connectivity.

In the temporal modeling stage, Temporal Multi-Head Self-Attention (T-MHSA) and GCN-DSR are similarly combined to jointly model cross-frame motion from both global temporal and dynamic structural perspectives. T-MHSA focuses on capturing long-range temporal patterns, whereas GCN-DSR models the dynamic evolution of motion-dependent relations over time.

Notably, GCN-DSR adopts a unified relation modeling mechanism across both spatial and temporal dimensions, enabling consistent learning of adaptive dependencies under varying motion states. This decouples dynamic relation construction from feature propagation, allowing adaptive dependency learning and structural feature aggregation to be optimized separately.

\textbf{Residual Skeletal Recalibration (RSRM).} 
To stabilize dynamically learned structural representations during cross-stream interaction, a Residual Skeletal Recalibration Module (RSRM) is introduced between spatial and temporal modeling stages. RSRM performs lightweight structure-guided feature recalibration in a residual manner, adaptively modulating joint-aware feature responses during feature propagation. This mechanism helps suppress unstable responses caused by noisy 2D keypoint observations and improves the consistency of motion-dependent structural representations across different modeling streams.

\textbf{Hierarchical Relation-aware Fusion.}  
To better exploit dynamically learned skeletal relations, a hierarchical relation-aware fusion strategy is designed to perform adaptive integration at both intra-block and cross-stream levels.

\textit{Intra-block Gated Fusion.} Within each DSR Hybrid Block, a learnable scalar gating parameter is introduced to adaptively weight the outputs of Multi-Head Self-Attention (MHSA) and the GCN-DSR branch. This establishes a balance between global semantic modeling and dynamic structural relation modeling. The mechanism enables the model to automatically adjust the contribution of each modeling branch according to motion-dependent feature characteristics.

\textit{Cross-stream Dynamic Channel Fusion (DCF).} Within each dual-stream DSR Hybrid Block, a Dynamic Channel Fusion (DCF) mechanism is employed to adaptively integrate features from different modeling paths. Specifically, DCF first applies global average pooling to extract channel-wise statistics from both branches and concatenates them to form a cross-stream semantic representation. A shared multilayer perceptron is then used to model the fused features and generate channel-wise weights for both branches.

Unlike traditional attention mechanisms that operate on individual branches, the channel weights in DCF are jointly determined by features from both branches, enabling cross-stream collaborative modulation. In addition, a Sigmoid activation function is applied to produce non-competitive channel weights, allowing the model to simultaneously preserve complementary information from multiple branches across different channels.

Finally, channel-wise reweighting is applied to fuse the features from both branches, achieving fine-grained dynamic feature integration. This mechanism effectively alleviates feature distribution discrepancies caused by different modeling orders and improves the consistency of dynamically learned structural representations across different modeling streams.

\textbf{Loss Function and Output Prediction.}
The model is trained in an end-to-end manner. The final feature representation, denoted as $F_N$, is projected into the 3D pose space through a fully connected layer followed by a $Tanh$ activation function. The loss function consists of a position loss ($L_{pos}$) and a velocity loss ($L_{vel}$), which jointly optimize joint coordinate accuracy and temporal smoothness. This ensures that the predicted 3D poses are both kinematically plausible and temporally consistent.

\subsection{Dynamic Skeletal Relation Modeling}

\begin{figure*}[t]
  \centering
  \includegraphics[width=\textwidth]{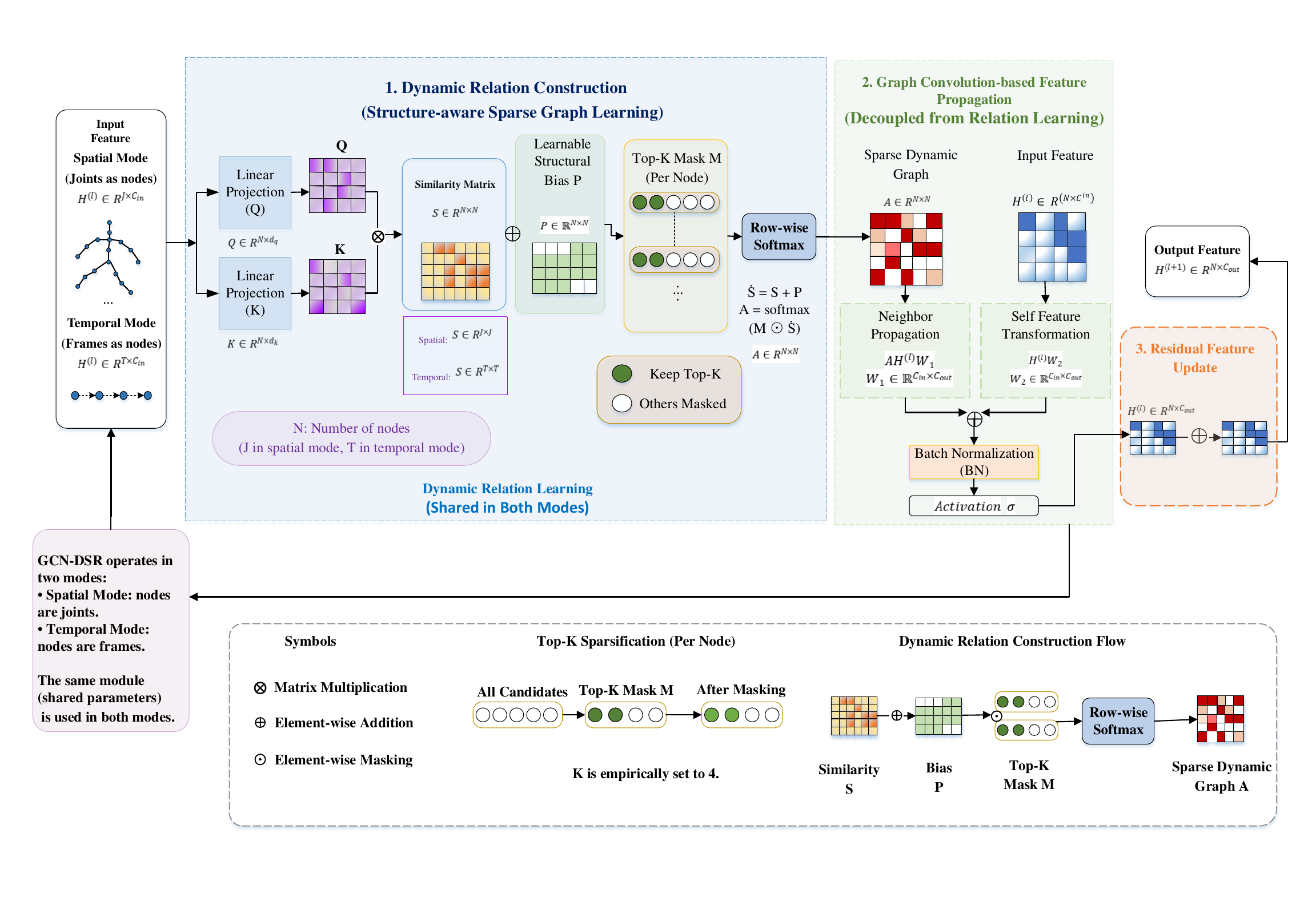}
  \caption{Overview of the proposed GCN-DSR module.
The framework first constructs sparse dynamic graph topologies through dynamic relation learning,
and then performs graph convolution-based feature propagation in a decoupled manner.
The same module is shared in both spatial and temporal modes.}
  \label{fig:gcndsr}
\end{figure*}

In 3D human pose estimation, the dependencies among human joints are highly motion-dependent and exhibit diverse structural patterns under different action states. However, traditional Graph Convolutional Network (GCN)-based methods typically rely on predefined skeletal topologies or static adjacency matrices, which are insufficient for modeling dynamically evolving spatial and temporal dependencies. Such fixed graph structures limit the representation capability of the network when handling complex human motions.

To address this issue, we propose a Dynamic Skeletal Relation Graph Convolution module (GCN-DSR) for adaptive modeling of motion-dependent joint relationships. 
The proposed module constructs dynamic relation graphs in a data-driven manner and performs graph-based feature propagation accordingly, enabling adaptive structural representation learning beyond fixed skeletal topology.

Given the input feature sequence 
\[
X \in \mathbb{R}^{B \times T \times J \times C},
\]
where \(B\), \(T\), \(J\), and \(C\) denote the batch size, temporal length, number of joints, and feature dimension, respectively, the proposed GCN-DSR operates in two decoupled stages: dynamic relation construction and graph-based feature propagation.

Specifically, the first stage is responsible for estimating dynamic graph topology from input features, while the second stage performs feature aggregation via graph convolution. 
The two stages are explicitly separated, ensuring that relation learning focuses solely on topology estimation without interfering with feature propagation.

\textbf{Unified Spatial-Temporal Relation Modeling.}
The proposed GCN-DSR performs unified spatial-temporal relation modeling via two symmetric operating modes, depending on whether joints or frames are treated as graph nodes.

In the \textit{spatial mode}, the input feature sequence is reshaped as
\[
\mathbb{R}^{(B \cdot T) \times J \times C},
\]
where joints are treated as graph nodes to model intra-frame structural dependencies.

In the \textit{temporal mode}, the input feature sequence is transposed and reshaped as
\[
\mathbb{R}^{(B \cdot J) \times T \times C},
\]
where frames are treated as graph nodes to capture inter-frame motion dependencies.

Both modes share the same dynamic relation construction mechanism, with the only difference being the node dimension, i.e., \(J\) for spatial modeling and \(T\) for temporal modeling.

\textbf{Dynamic Relation Construction.}
Given the input features, dynamic relations are estimated through query-key feature projection:
\[
Q = XW_Q, \quad K = XW_K,
\]
where \(W_Q\) and \(W_K\) are learnable projection matrices.

The similarity matrix is computed using scaled dot-product similarity:
\[
S = \frac{QK^T}{\sqrt{d_q}},
\]
where \(d_q\) denotes the feature dimension of the projected query and key representations.

Specifically, in spatial mode,
\[
S_s \in \mathbb{R}^{J \times J}
\]
models dynamic inter-joint dependencies, while in temporal mode,
\[
S_t \in \mathbb{R}^{T \times T}
\]
captures motion-dependent correlations across frames.

To improve structural stability during relation learning, a learnable structural bias matrix is further introduced:
\[
\tilde{S} = S + P,
\]
where
\[
P \in \mathbb{R}^{N \times N}
\]
denotes a learnable structural bias matrix, with \(N = J\) in spatial mode and \(N = T\) in temporal mode.

Both spatial and temporal bias matrices are initialized to near-zero values and optimized end-to-end during training, enabling adaptive relation refinement without imposing hard predefined connectivity constraints.

Directly modeling dense relations may introduce redundant connections and increase computational complexity. Therefore, a Top-K sparsification strategy is employed to retain only the most informative relations. Specifically, for each node, only the Top-K neighboring nodes with the highest similarity scores are preserved:
\[
M_{ij} =
\begin{cases}
1, & \text{if } j \in \text{Top-K}(i), \\
-\infty, & \text{otherwise}.
\end{cases}
\]

In this work, \(K\) is empirically set to 4, which provides a favorable trade-off between graph sparsity and relation representation capability according to the ablation study. 

The final dynamic adjacency matrix is obtained via Softmax normalization:
\[
A = \text{softmax}(M \odot \tilde{S}).
\]

The learnable similarity matrix and Top-K sparsification jointly determine the final dynamic graph structure, allowing the model to adaptively preserve important spatial and temporal dependencies while suppressing noisy correlations.

\textbf{Graph-based Feature Propagation.}
For notation consistency in the following formulations, 
we denote the reshaped node feature as 
$H^{(l)} \in \mathbb{R}^{N \times C}$, 
where $N = J$ in spatial mode and $N = T$ in temporal mode.

After obtaining the dynamic adjacency matrix \(A\), graph convolution is performed for feature propagation:
\[
H^{l+1}
=
H^l
+
\sigma
\left(
BN
\left(
AH^lW_1 + H^lW_2
\right)
\right),
\]
where \(H^l\) and \(H^{l+1}\) denote the input and output features of the \(l\)-th layer, respectively; \(W_1\) and \(W_2\) are learnable projection matrices for neighborhood propagation and self-feature transformation; \(BN\) denotes batch normalization; and \(\sigma\) denotes nonlinear activation.

Unlike traditional Graph Attention Networks (GAT), the attention mechanism in the proposed method is only used to dynamically estimate relation structures, while feature propagation is performed through graph convolution operations. Therefore, relation modeling and feature propagation are explicitly decoupled.

After graph propagation, the output features are reshaped back to the original spatiotemporal representation:
\[
\mathbb{R}^{B \times T \times J \times C},
\]
ensuring consistent feature dimensions across different modeling stages.

This decoupled dynamic relation modeling strategy enables the proposed GCN-DSR to flexibly capture motion-dependent structural variations while maintaining stable graph-based feature propagation.

\subsection{Skeletal Feature Recalibration}

In 3D human pose estimation, the human skeleton follows explicit topological constraints. Although dynamic skeletal relation modeling enables adaptive structural representation learning, feature distributions produced by different modeling streams may still exhibit inconsistency during branch interaction.

In the dual-branch structure (S-T / T-S), the learned feature distributions from different branches are not identical. Direct feature fusion may introduce structural distortion. Therefore, introducing a structural recalibration mechanism before branch interaction is important to align features and improve the stability and effectiveness of subsequent fusion.

\begin{figure}[pos=htbp]
  \centering
  \includegraphics[width=0.95\linewidth, height=0.4\textheight, keepaspectratio]{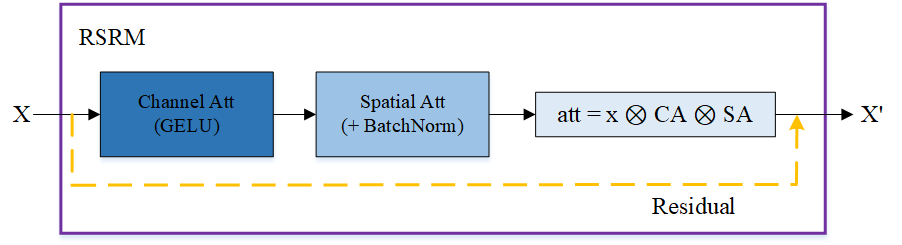}
  \caption{Architecture of the proposed RSRM module.}
  \label{fig:rsrm}
\end{figure}

Based on this analysis, we introduce a Residual Skeletal Recalibration Module (RSRM) at the branch interaction stage. This module enhances structural consistency before feature fusion, thereby improving the geometric plausibility and semantic representation of the human skeleton. The overall structure of this module is shown in Fig. \ref{fig:rsrm}.

The module is designed to be structure-aware, lightweight, and stable through residual learning, enabling effective feature recalibration without introducing significant computational overhead.

\textbf{Structure-aware Feature Modeling.}
Let the input feature be defined as: $\mathbf{X} \in \mathbb{R}^{B \times T \times J \times C}$, where $B$, $T$, $J$, and $C$ denote the batch size, temporal length, number of joints, and channel dimension, respectively. First, the temporal dimension is compressed to obtain joint-level structural representations: $\mathbf{X}_s = \text{Pool}_t(\mathbf{X})$, where $\text{Pool}_t(\cdot)$ denotes a statistical operation along the temporal dimension (e.g., average pooling). Subsequently, a skeletal topology adjacency matrix is introduced: $\mathbf{A} \in \mathbb{R}^{J \times J}$, and structural information is modeled through graph convolution:
\[
\mathbf{S} = \sigma(\mathbf{A} \mathbf{X}_s \mathbf{W}),
\]
where $\mathbf{W}$ is a learnable weight matrix and $\sigma(\cdot)$ denotes a nonlinear activation function (e.g., GELU). This process models local dependencies among joints and extracts a structure-aware representation $\mathbf{S}$, which serves as an auxiliary skeletal prior for preserving topology-consistent feature responses during the subsequent recalibration process.

\textbf{Structure-aware Residual Recalibration.}
The proposed RSRM performs structure-aware dual-branch residual recalibration through channel attention and spatial attention mechanisms.

For channel-wise recalibration, global average pooling and max pooling are first applied to aggregate global contextual information:
\[
\mathbf{A}_c
=
\sigma
(
\text{MLP}(\text{AvgPool}(\mathbf{X}))
+
\text{MLP}(\text{MaxPool}(\mathbf{X}))
),
\]
where \(\mathbf{A}_c\) denotes the channel attention weights and \(\sigma(\cdot)\) represents the sigmoid activation function.

The extracted structure-aware representation serves as an auxiliary skeletal prior, providing structural guidance for the subsequent residual recalibration process.

Subsequently, spatial attention is estimated based on aggregated channel statistics:
\[
\mathbf{A}_s
=
\sigma
(
f^{7\times7}
(
[\text{Avg}(\mathbf{X});\text{Max}(\mathbf{X})]
)
),
\]
where \(f^{7\times7}\) denotes a 2D convolution applied on the reshaped temporal–joint feature map, i.e., \(\mathbf{X} \in \mathbb{R}^{B \times T \times J \times C}\) is first permuted to \(\mathbb{R}^{B \times C \times T \times J}\), and the convolution is performed over the \((T \times J)\) domain to model long-range spatiotemporal dependencies.

The final recalibrated feature is obtained through residual attention fusion:
\[
\mathbf{X}_{\text{out}}
=
\mathbf{X}
+
(
\mathbf{X}
\odot
\mathbf{A}_c
\odot
\mathbf{A}_s
),
\]
where \(\odot\) denotes element-wise multiplication.

Unlike generic channel-spatial attention mechanisms that solely rely on feature statistics, the proposed RSRM introduces skeletal structural priors into the recalibration process, leading to more structure-consistent feature refinement under complex motion dynamics.

\subsection{Hierarchical Relation-aware Fusion}

Although the proposed DSR framework enables dynamic skeletal relation learning, the resulting features from different modeling branches exhibit heterogeneous semantic characteristics. Direct fusion through fixed weighting or simple summation may therefore lead to feature redundancy and suboptimal utilization of complementary information. 

To address this issue, we propose a hierarchical relation-aware fusion strategy that performs progressive and adaptive integration of multi-source features at two levels: intra-block complementary modeling and cross-stream adaptive reweighting.

\textbf{Intra-block Fusion.}
Within each Spatial/Temporal DSR Block, a dual-branch parallel structure is adopted, consisting of a Multi-Head Self-Attention (MHSA) branch and a graph-based dynamic relation modeling branch (GCN-DSR). The MHSA branch models long-range dependencies among joints through a global attention mechanism, enabling the capture of global contextual information. In contrast, the GCN-DSR branch explicitly models motion-dependent skeletal relations through dynamic graph construction, enabling adaptive structural representation beyond fixed skeletal connectivity.

Given the input feature $\mathbf{X} \in \mathbb{R}^{B \times T \times J \times C}$, where $B$, $T$, $J$, and $C$ denote the batch size, number of frames, number of joints, and channel dimension, respectively. The input is first normalized and then fed into the two branches separately:
\[
\mathbf{X}_a = \text{MHSA}(\text{LN}(\mathbf{X})), \quad \mathbf{X}_g = \text{GCN-DSR}(\text{LN}(\mathbf{X}))
\]

To enable adaptive fusion of the two types of features, a learnable scalar parameter $\lambda$ is introduced and constrained by a Sigmoid function:
\[
\alpha = \sigma(\lambda), \quad \alpha \in (0,1)
\]
The fused feature representation within the block is then given by:
\[
\mathbf{X}_{ib} = \alpha \cdot \mathbf{X}_a + (1 - \alpha) \cdot \mathbf{X}_g
\]
Here, $\alpha$ controls the relative contribution of the MHSA branch and the GCN-DSR branch during feature fusion.
The fused features are further passed through a feed-forward network (MLP) for nonlinear transformation, combined with residual connections and a LayerScale mechanism to enhance training stability. Through this gating mechanism, the model can dynamically adjust the contribution of the MHSA and GCN-DSR branches according to different inputs, thereby fully exploiting the complementary strengths of the two modeling approaches.

\textbf{Cross-stream Fusion.}
At a higher level, the model adopts a parallel dual-stream structure (e.g., ST and TS pathways) to model spatiotemporal features from different perspectives. 
Although collaborative modeling is performed within each block, discrepancies in semantic representation and feature distribution still exist across streams. 
Direct averaging-based fusion cannot fully exploit the advantages of the dual-stream design.

To address this, we propose a \textit{Dynamic Channel Fusion (DCF)} module that performs channel-wise adaptive reweighting of features from different streams. 
Let the output features of the Spatial-First and Temporal-First streams be denoted as
\[
\mathbf{X}_{ST}, \mathbf{X}_{TS} \in \mathbb{R}^{B \times T \times J \times C}.
\]
First, global average pooling is applied along the temporal and joint dimensions to extract channel-wise statistical representations:
\[
z_{ST} = \text{GAP}(\mathbf{X}_{ST}), \quad z_{TS} = \text{GAP}(\mathbf{X}_{TS}),
\]
where $z_{ST}, z_{TS} \in \mathbb{R}^{B \times C}$.

The two representations are then concatenated to obtain a joint channel descriptor:
\[
z = [z_{ST}, z_{TS}] \in \mathbb{R}^{B \times 2C}.
\]
This joint descriptor captures complementary channel statistics from the two streams for subsequent adaptive reweighting. This feature vector is passed through a lightweight MLP to generate channel-wise weights:
\[
w = \phi(z) \in \mathbb{R}^{B \times 2C},
\]
where $\phi(\cdot)$ denotes a nonlinear mapping function composed of two fully-connected layers.

The output is further split into two groups of weights:
\[
w_{ST}, w_{TS} \in \mathbb{R}^{B \times C},
\]
and normalized using a Sigmoid function:
\[
\hat{w}_{ST} = \sigma(w_{ST}), \quad \hat{w}_{TS} = \sigma(w_{TS}).
\]
Finally, the weights are expanded to the original feature dimensions and applied for channel-wise weighted fusion:
\[\mathbf{X}_{cs} =
\hat{w}_{ST} \odot \mathbf{X}_{ST}
+
\hat{w}_{TS} \odot \mathbf{X}_{TS}\]

where $\odot$ denotes channel-wise multiplication. This operation enables channel-wise adaptive propagation of complementary semantic information from different streams.

Unlike conventional fusion strategies that employ static weighting or direct feature aggregation, the proposed DCF module jointly determines channel-wise importance from both streams, enabling adaptive cross-stream semantic interaction.

This hierarchical fusion strategy enables adaptive integration of multi-source features at different levels. At the intra-block level, the gating mechanism facilitates dynamic collaboration between MHSA and GCN-DSR, allowing the model to balance global semantic modeling and local structural modeling. At the cross-stream level, the DCF module performs channel-wise reweighting to align feature distributions and enhance complementary information across streams. 

Overall, the proposed hierarchical relation-aware fusion strategy improves feature complementarity, structural consistency, and cross-stream semantic interaction, resulting in more discriminative spatiotemporal representations for dynamic human pose estimation.

\section{Experiments}

\subsection{Datasets and Metrics}

We evaluate the proposed DSR-Net on two widely used benchmarks, Human3.6M~\cite{ionescu_human3_2013} and MPI-INF-3DHP~\cite{mehta_monocular_2017}, to assess both in-domain performance and cross-dataset generalization. In addition, qualitative visualization experiments are conducted on the 3DPW \cite{3dpw2018} dataset to further evaluate the robustness and generalization capability of the proposed method in challenging in-the-wild scenarios.

Human3.6M is a widely used benchmark dataset for 3D human pose estimation, consisting of large-scale indoor motion capture data with diverse actions and subjects. Following the standard Protocol \#1, subjects S1, S5, S6, S7, and S8 are used for training, while S9 and S11 are used for testing. Ground-truth 2D keypoints are used as input, and the human body is represented by 17 joints.

On Human3.6M, we report Mean Per Joint Position Error (MPJPE, also denoted as P1) as the primary metric~\cite{hossain_exploiting_2018,lee_propagating_2018}:
\[
\text{MPJPE} = \frac{1}{N} \sum_{i=1}^{N} \| \hat{p}_i - p_i \|_2
\]
where $N$ denotes the number of joints, and $\hat{p}_i$ and $p_i$ are the predicted and ground-truth 3D joint coordinates, respectively.

We also report Procrustes-aligned MPJPE (P-MPJPE, or P2), where a rigid alignment is applied before error computation~\cite{lee_propagating_2018}.

MPI-INF-3DHP~\cite{mehta_monocular_2017} is used to evaluate cross-dataset generalization under more diverse conditions, including indoor and outdoor scenes. It contains multiple subjects and complex motions with an independent test set. 

Following prior work, we report MPJPE, Percentage of Correct Keypoints (PCK) with a threshold of 150 mm, and Area Under the Curve (AUC).

To further evaluate the qualitative performance of DSR-Net in real-world environments, additional visualization experiments are conducted on the 3DPW dataset. Compared with indoor benchmark datasets, 3DPW contains outdoor scenes with viewpoint variations, motion diversity, and partial occlusions, providing a more challenging evaluation setting for qualitative analysis. Since the focus of this experiment is qualitative visualization rather than quantitative benchmarking, no additional evaluation metrics are reported for 3DPW.

\subsection{Implementation Details}

DSR-Net is built upon MixTGFormer~\cite{duan_dual-stream_2026}, which serves as the baseline architecture in this work. The proposed model retains the original dual-stream Transformer framework while introducing the Dynamic Skeletal Relation Graph Convolution module (GCN-DSR), the Residual Skeletal Recalibration Module (RSRM), and the hierarchical relation-aware fusion strategy. Except for the explicitly introduced modules, all network settings remain identical to the baseline configuration to ensure fair comparison.

All experiments are implemented using the PyTorch deep learning framework and conducted on a single NVIDIA GeForce RTX 4090 GPU. The model is trained using the AdamW optimizer with an initial learning rate of $5\times10^{-4}$ and a weight decay of 0.01. An exponential decay schedule with a decay factor of 0.99 is adopted during training. Due to the long temporal sequence length (243 frames) and the additional computational overhead introduced by dynamic graph modeling, the batch size is set to 4 to satisfy GPU memory constraints. All models are trained for 60 epochs without early stopping, and the random seed is fixed to 42 to ensure reproducibility.

During preprocessing, the input 2D keypoint sequences are normalized with respect to the root joint (pelvis), and the coordinates are scaled to a unified range for improved numerical stability. Horizontal flipping is adopted as the primary data augmentation strategy during both training and testing. The preprocessing and evaluation protocols generally follow standard settings commonly adopted in previous 3D human pose estimation studies~\cite{gong_poseaug_2021,zhao_semantic_2019,zou_modulated_2021,xu_deep_2020}.

For temporal modeling, input sequences with a length of 243 frames are used. During training, long motion sequences are segmented using a sliding-window strategy, while the inference protocol remains consistent with the baseline model. MPJPE is adopted as the primary evaluation metric and optimization objective throughout all experiments.

On the Human3.6M dataset, we follow the standard evaluation protocol by using both the ground-truth 2D annotations and the 2D keypoints generated by the Stacked Hourglass detector~\cite{newell_stacked_2016} as input. On the MPI-INF-3DHP dataset, the ground-truth 2D keypoints provided by the dataset are directly used without relying on external 2D pose detectors.

Unless otherwise specified, all ablation studies are conducted based on the reproduced MixTGFormer baseline under identical training and evaluation settings. For controlled analysis, only one target component is modified at a time while all remaining settings are kept unchanged.

\subsection{Results on Human3.6M}

\subsubsection{Overall Performance Comparison}

To evaluate the effectiveness of the proposed DSR-Net, we conduct quantitative comparisons on the Human3.6M dataset against the baseline model and several representative methods under controlled indoor environments. To ensure fair comparison, all experiments are performed under identical training settings, using the same 2D detector inputs and evaluation protocol whenever possible.

Table~\ref{tab:h36m} summarizes the quantitative results on Human3.6M. Under the same backbone architecture and training strategy, the proposed DSR-Net reduces the MPJPE from 39.1 mm to 37.9 mm, demonstrating consistent performance improvement over the baseline model.

\begin{figure*}[pos=htbp]
    \centering
    \includegraphics[width=0.6\textwidth, keepaspectratio]{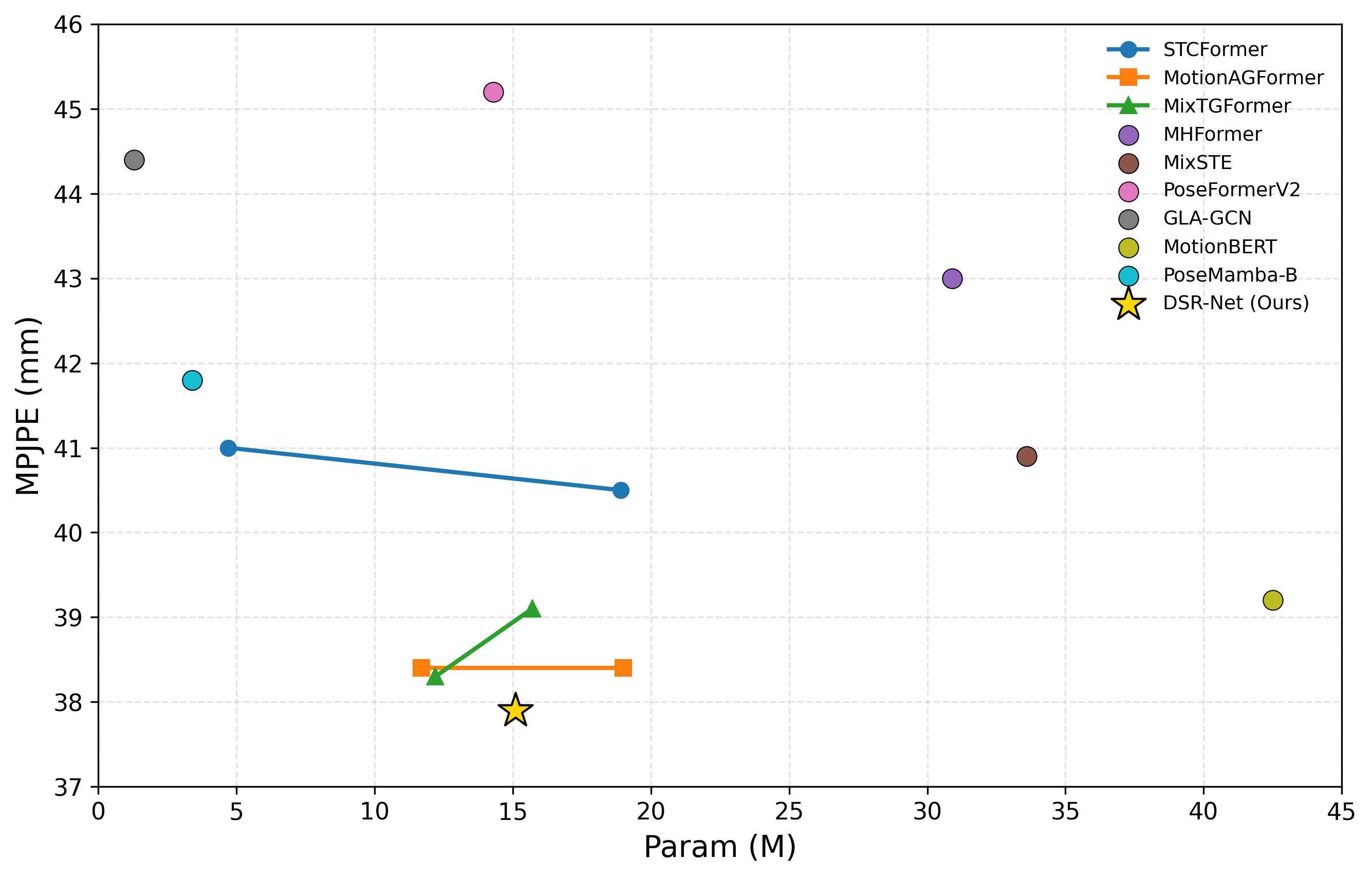}
    \caption{
    Comparison with representative 3D human pose estimation methods on the Human3.6M dataset. MPJPE denotes the mean per-joint position error (lower is better), while Param indicates the number of model parameters.
    }
    \label{fig:tradeoff}
\end{figure*}

\begin{table}[htbp]
  \centering
  \caption{Quantitative comparison results on Human3.6M.
  T: Number of input frames; P1: MPJPE error (mm); P2: P-MPJPE error (mm);
  Param and MACs indicate the number of parameters and computational complexity, respectively.
  The best and second-best results are shown in bold and underlined, respectively.}
  \label{tab:h36m}
  \setlength{\tabcolsep}{3pt}
  \resizebox{\linewidth}{!}{
  \begin{tabular}{lccccc}
    \toprule
    Method & T & Param (M)& MACs (G)& P1$\downarrow$ & P2$\downarrow$ \\
    \midrule
    MHFormer\cite{li_mhformer_2021} & 351 & 30.9& 7.0& 43.0 & 34.4 \\
    MixSTE\cite{zhang_mixste_2022} & 243 & 33.6& 139.0& 40.9 & 32.6 \\
    STCFormer\cite{tang_3d_2023}& 243 & 4.7& 19.6& 41.0 & 32.0 \\
    STCFormer-L\cite{tang_3d_2023}& 243 & 18.9& 78.2& 40.5 & \underline{31.8} \\
    PoseFormerV2\cite{zhao_poseformerv2_2023}& 243 & 14.3& 0.5& 45.2 & 35.6 \\
    GLA-GCN\cite{yu_gla-gcn_2023} & 243 & 1.3& 1.5& 44.4 & 34.8 \\
    MotionBERT\cite{zhu_motionbert_2023} & 243 & 42.5& 174.7& 39.2 & 32.9 \\
    STGFormer\cite{liu_stgformer_2026}& 243 & 31.9& - & 40.3 & \textbf{31.7 }\\
    STAPFormer\cite{zhang_stapformer_2024}& 243 & - & - & 38.2& 32.1 \\
    PoseMamba-B\cite{huang_posemamba_2025} & 243 & 3.4& 13.9& 41.8 & 35.0 \\
    MotionAGFormer-B\cite{mehraban_motionagformer_2024} & 243 & 11.7& 48.3& 38.4 & 32.6 \\
    MotionAGFormer-L\cite{mehraban_motionagformer_2024} & 243 & 19.0& 78.3& 38.4 & 32.5 \\
    DSTFormer\cite{zhang_dstformer_2024} & 243 & 12.0& 48.3& \underline{37.9}& 32.2 \\
    MixTGFormer$^\dagger$\cite{duan_dual-stream_2026} & 243 & 15.7& 63.4& 39.1 & 32.8 \\
    DSR-Net (Ours) & 243 & 15.1& 60.9& \textbf{37.9} & 31.9 \\
    \bottomrule
  \end{tabular}
  }
  \vspace{1mm}
  \begin{flushleft}
  \footnotesize{
  $^\dagger$ denotes our reproduced result under identical training and evaluation settings.  The reproduced performance is slightly lower than the result reported in the official paper, which may be attributed to differences in implementation details and training environment. All ablation studies and model comparisons are conducted based on this reproduced baseline to ensure fair evaluation.
  }
  \end{flushleft}
\end{table}

Fig.~\ref{fig:tradeoff} further illustrates the relationship between model complexity and pose estimation accuracy. Compared with representative methods, the proposed DSR-Net achieves competitive performance while maintaining a relatively moderate parameter size, demonstrating a favorable balance between model capacity and estimation accuracy.

It is worth noting that the reproduced baseline performance in our implementation is slightly lower than the results reported in the original paper. This discrepancy is commonly observed when reproducing existing methods due to differences in implementation details, training environments, and randomness. Under a fixed random seed and strictly controlled training configuration, the reproduced baseline achieves stable performance, thus serving as a reliable reference for relative comparison. Under the same experimental settings, DSR-Net consistently achieves performance improvements over the baseline, demonstrating stable gains and indicating the effectiveness of the proposed design. 

For some methods, computational complexity (MACs) is not publicly available and is therefore denoted as “–” in Table~\ref{tab:h36m}. Although certain methods (e.g., DSTFormer) achieve comparable MPJPE performance under different architectural configurations, the two methods adopt fundamentally different modeling paradigms. DSTFormer mainly relies on pure Transformer-based dependency modeling, whereas DSR-Net explicitly incorporates dynamic skeletal relation learning and structure-aware feature propagation through GCN-DSR and RSRM. The additional parameters primarily arise from these structural modeling components rather than redundant network expansion.

Moreover, while the two methods obtain comparable MPJPE results, DSR-Net achieves a lower P-MPJPE (31.9 mm vs. 32.2 mm), indicating improved structural consistency after pose alignment. This suggests that the proposed framework is more effective in capturing underlying skeletal relations and structure-aware pose representations.

Combined with the ablation studies, the observed performance improvements can be attributed to several key components. The proposed dynamic skeletal relation modeling enhances the flexibility of inter-joint dependency learning, enabling more accurate structural representation of the human skeleton. In addition, the Residual Skeletal Recalibration Module (RSRM) improves the consistency of skeletal feature representations, while the dynamic fusion strategy facilitates adaptive integration of complementary information across different feature branches.

Overall, the experimental results demonstrate that DSR-Net effectively improves pose estimation accuracy while maintaining competitive model complexity, validating the effectiveness of dynamic skeletal relation modeling for 3D human pose estimation.

\subsubsection{Action-wise Performance Comparison}

To further analyze the model performance across different action categories, we report the MPJPE results of all methods on individual actions in the Human3.6M dataset. The results are presented in Table \ref{tab:h36m_actions}. It can be shown that the proposed method achieves the best or second-best performance on most actions, with notable improvements on representative actions such as Walking, Eating, and Discussion.
In contrast, for actions involving large pose variations and self-occlusion, such as Sitting, the performance gap among different methods is relatively small, indicating that these scenarios 
remain challenging. Overall, the action-wise results further demonstrate the stability and generalization capability of the proposed method across diverse motion patterns.

\begin{table*}[htbp]
  \centering
 \caption{MPJPE results (mm) on the Human3.6M dataset under Protocol \#1 (P1) and Protocol \#2 (P2). All methods use SH-detected 2D keypoints for a fair comparison.}
  \label{tab:h36m_actions}
  \setlength{\tabcolsep}{2.5pt} 
  \renewcommand{\arraystretch}{1.05}
  \begin{tabular}{lcccccccccccccccccc}
    \toprule
    Method & T & Dir. & Disc. & Eat. & Greet & Phone & Photo & Pose & Pur. & Sit & SitD. & Smoke & Wait & WalkD. & Walk & WalkT. & Avg. \\
    \midrule
    \multicolumn{18}{l}{MPJPE (P1)} \\
    SemGCN\cite{zhao_semantic_2019} & 1 & 48.2 & 60.8 & 51.8 & 64.0 & 64.6 & 53.6 & 51.1 & 67.4 & 88.7 & 57.7 & 73.2 & 65.6 & 48.9 & 64.8 & 51.9 & 60.8 \\
    EHGFormer\cite{zheng_ehgformer_2025} & 243 & \textbf{34.1} & \textbf{38.0} & 38.2 & \textbf{32.4} & \textbf{40.6} & \underline{47.8} & 37.3 & \textbf{34.3} & 50.2 & 56.1 & \textbf{39.4} & \underline{36.7} & \textbf{26.3} & 35.8 & \textbf{25.8} & \underline{38.2} \\
    MotionBERT\cite{zhu_motionbert_2023} & 243 & 36.3 & 38.7 & 38.6 & 33.6 & 42.1 & 50.1 & \textbf{36.2} & 35.7 & \underline{50.1} & 56.6 & 41.3 & 37.4 & 37.7 & \underline{25.6 }& 26.5 & 39.2 \\
    MotionAGFormer\cite{mehraban_motionagformer_2024} & 243 & 36.4 & \underline{38.4} & \textbf{36.8} & \underline{32.9} & 40.9 & 48.5 & \underline{36.6} & 34.6 & 51.7 & \underline{52.8} & 41.0 & \textbf{36.4} & 36.5 & 26.7 & 27.0 & 38.4 \\
    DSR-Net (Ours) & 243 & \underline{36.0} & \underline{38.4} & \underline{36.9} & 33.2 & \underline{40.7} & \textbf{47.2} & \textbf{36.2} & \underline{34.6} & \textbf{48.7} & \textbf{51.5} & \underline{40.5} & \underline{36.7} & \underline{36.0} & \textbf{25.5} & \underline{25.9} & \textbf{37.9} \\
    \midrule
    \multicolumn{18}{l}{P-MPJPE (P2)} \\
    EHGFormer\cite{zheng_ehgformer_2025} & 243 & \textbf{29.1} & \textbf{31.9} & 32.4 & \textbf{27.5} & \textbf{32.9} & \underline{38.1} & 30.5 & \textbf{29.5} & 43.6 & 49.7 & \textbf{34.6} & 30.2 & \underline{22.3} & 30.7 & \underline{22.5} & \underline{32.4} \\
    MotionBERT\cite{zhu_motionbert_2023} & 243 & 30.8 & 32.8 & 32.4 & 28.7 & 34.3 & 38.9 & \underline{30.1} & 30.0 & \underline{42.5} & 49.7 & 36.0 & 30.8 & \textbf{22.0} & 31.7 & 23.0 & 32.9 \\
    MotionAGFormer\cite{mehraban_motionagformer_2024} & 243 & 30.6 & 32.6 & \underline{32.2} & 28.2 & 33.8 & 38.6 & 30.5 & 29.9 & 43.3 & \underline{47.0} & 35.2 & \textbf{29.8} & 31.4 & \underline{22.7} & 23.5 & 32.6 \\
DSR-Net (Ours) & 243 & \underline{30.3} & \underline{32.5} & \textbf{31.9} & \underline{28.1} & \underline{33.5} & \textbf{37.3} & \textbf{29.7} & \underline{29.8} & \textbf{40.9} & \textbf{45.1} & \underline{35.1} & \underline{30.1} & 30.7 & \textbf{21.5} & \textbf{22.2} & \textbf{31.9} \\
    \bottomrule
  \end{tabular}
\end{table*}

\subsection{Results on MPI-INF-3DHP}

\begin{table}[htbp]
  \centering
  \caption{Quantitative comparison with existing methods on the MPI-INF-3DHP dataset.
  PCK and AUC denote Percentage of Correct Keypoints and Area Under the Curve, respectively.
  ``T'' represents the number of input frames.}
  \label{tab:mpi}
  \setlength{\tabcolsep}{4pt}
  \begin{tabular}{lcccc}
    \toprule
    Method & T & PCK$\uparrow$ & AUC$\uparrow$ & MPJPE$\downarrow$ \\
    \midrule
    STCFormer\cite{tang_3d_2023} & 81 & 98.7 & 83.9 & 23.1 \\
    PoseFormerV2\cite{zhao_poseformerv2_2023} & 81 & 97.9 & 78.8 & 27.8 \\
    GLA-GCN\cite{yu_gla-gcn_2023} & 81 & 98.5 & 79.1 & 27.7 \\
    DiffPose\cite{gong_diffpose_2023} & 81 & 98.0 & 75.9 & 29.1 \\
    CGFusionFormer\cite{lu_cgfusionformer_2025} & 81 & 97.9 & 78.5 & 27.2 \\
    STAPFormer\cite{zhang_stapformer_2024} & 81 & \underline{98.8} & 83.9 & 20.5 \\
    KTPFormer\cite{peng_ktpformer_2024} & 81 & \textbf{98.9} & 85.9 & 16.7 \\
    MotionAGFormer\cite{mehraban_motionagformer_2024} & 81 & 98.3 & 84.2 & 18.2 \\
    MixTGFormer-s\cite{duan_dual-stream_2026} & 81 & 98.3 & 84.7 & 18.6 \\
    MixTGFormer\cite{duan_dual-stream_2026} & 81 & 98.4 & \underline{86.9} & \underline{15.9} \\
   DSR-Net (Ours) & 81 & 98.5 & \textbf{87.8} & \textbf{14.9} \\
    \bottomrule
  \end{tabular}
\end{table}

On the MPI-INF-3DHP dataset, we evaluate the proposed DSR-Net under a cross-dataset generalization setting and compare it with representative state-of-the-art methods under identical experimental settings. All experiments are conducted with an input sequence length of 81 frames under the same training configuration. To ensure reproducibility, the random seed is fixed to 42 for all experiments. The quantitative results are summarized in Table~\ref{tab:mpi}.

From Table~\ref{tab:mpi}, the proposed DSR-Net achieves the best performance in terms of MPJPE and AUC. Specifically, it reduces the MPJPE from 15.9 mm to 14.9 mm under identical settings, demonstrating consistent improvement in cross-dataset scenarios.

In terms of PCK and AUC, DSR-Net achieves 98.5\% and 87.8\%, respectively, maintaining competitive accuracy while achieving the highest AUC among all compared methods. This indicates that the proposed method not only improves prediction accuracy but also enhances stability across different error thresholds.

The performance gains can be attributed to the proposed dynamic relation modeling and hierarchical fusion mechanisms. The dynamic skeletal relation modeling (DSR) enables the model to adaptively capture structural dependencies under varying motion patterns, which is particularly beneficial in cross-domain scenarios with diverse poses and environments. In addition, the hierarchical relation-aware fusion strategy facilitates effective integration of multi-branch features, improving the robustness and consistency of pose estimation.

Overall, these results demonstrate that DSR-Net generalizes well to unseen scenarios and achieves strong performance on the MPI-INF-3DHP benchmark.

\subsection{Ablation Studies}

To systematically evaluate the effectiveness of each proposed component, we conduct a series of ablation studies on the Human3.6M dataset under identical training and evaluation settings.

\subsubsection{Progressive Integration of Modules}

\begin{figure*}[pos=htbp]
    \centering
    \includegraphics[width=0.55\textwidth, keepaspectratio]{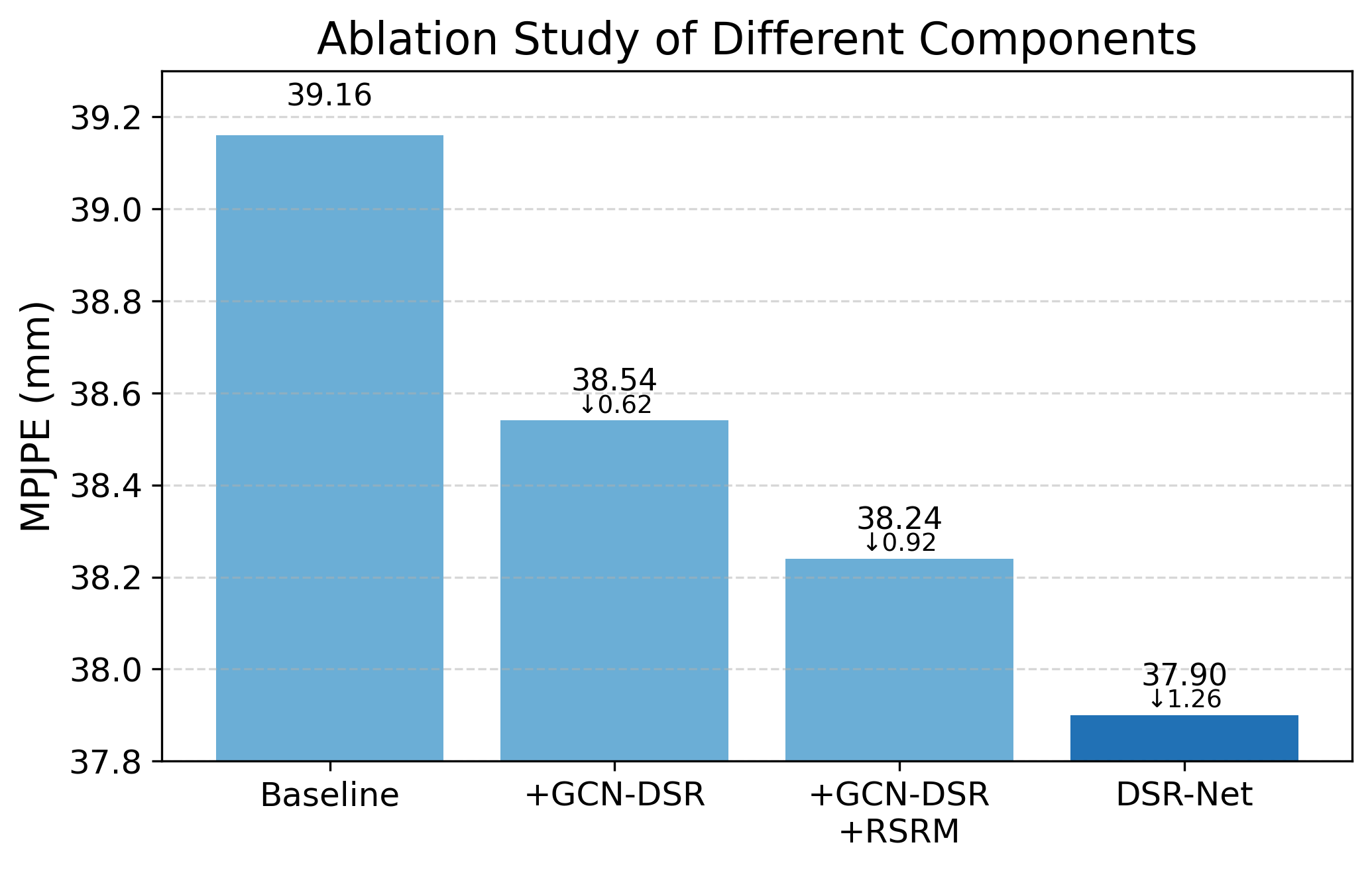}
    \caption{Ablation study of different components on the Human3.6M dataset. Each proposed module progressively improves the pose estimation performance, leading to a cumulative reduction of 1.26 mm in MPJPE.}
    \label{fig:ablation}
\end{figure*}

We first evaluate the cumulative contribution of each proposed component through progressive integration based on the reproduced MixTGFormer baseline. Specifically, the Dynamic Skeletal Relation Graph Convolution module (GCN-DSR), the Residual Skeletal Recalibration Module (RSRM), and the Hierarchical Relation-aware Fusion strategy (HRF) are sequentially integrated into the baseline framework. The corresponding results are reported in Table~\ref{tab:ablation}, while Fig.~\ref{fig:ablation} provides a visual comparison of the progressive performance improvements achieved by different components.

\begin{table}[htbp]
  \centering
  \caption{Progressive integration study of the proposed modules.}
  \label{tab:ablation}
  \setlength{\tabcolsep}{3pt}
  \begin{tabular}{ccccc}
    \toprule
    GCN & RSRM & HRF & MPJPE & $\Delta$ \\
    \midrule
    $\times$     & $\times$     & $\times$     & 39.16 & --   \\
    $\checkmark$ & $\times$     & $\times$     & 38.54 & 0.62 \\
    $\checkmark$ & $\checkmark$ & $\times$     & 38.24 & 0.92 \\
    $\checkmark$ & $\checkmark$ & $\checkmark$ & 37.90 & 1.26 \\
    \bottomrule
  \end{tabular}
\end{table}

As shown in Table~\ref{tab:ablation}, integrating GCN-DSR reduces the MPJPE from 39.16 mm to 38.54 mm, demonstrating the effectiveness of dynamic relation modeling for capturing inter-joint dependencies. Further incorporating RSRM decreases the MPJPE to 38.24 mm, indicating that structure-aware feature recalibration improves feature representation stability. After additionally introducing HRF, the model achieves the best performance of 37.90 mm, yielding an overall improvement of 1.26 mm over the baseline.

Overall, the progressive integration results indicate that the proposed components contribute complementary improvements to pose estimation performance.

Although the above experiments validate the overall effectiveness of the proposed modules, the progressive integration strategy may introduce interdependencies among different components. To further isolate and analyze the contribution of GCN-DSR, we conduct a dedicated ablation study, with the corresponding results presented as follows.

\subsubsection{Analysis of GCN-DSR}

To evaluate the effectiveness of the proposed GCN-DSR, we investigate different graph construction configurations, including static graph modeling, spatial dynamic relation modeling, and joint spatiotemporal dynamic relation modeling.

The baseline configuration employs a fixed skeletal graph topology. On this basis, dynamic spatial relations are adaptively constructed according to feature similarity between joints. Temporal constraints are further incorporated to regularize the dynamic graph construction process, thereby forming the complete GCN-DSR framework.

The experimental results are reported in Table~\ref{tab:graph_model}. When only spatial dynamic relation modeling is introduced, the performance slightly degrades. This suggests that dynamically constructed spatial relations are susceptible to noise without temporal constraints, leading to unstable graph structures.

\begin{table*}[htbp]
  \centering
  \caption{Analysis of different dynamic relation modeling configurations.}
  \label{tab:graph_model}
  \setlength{\tabcolsep}{3pt}
  
  \begin{tabular}{cccc}
    \toprule
    Dynamic Spatial Relation & Temporal Constraint & Graph Structure & MPJPE \\
    \midrule
    $\times$     & $\times$     & Static Graph & 39.16 \\
    $\checkmark$ & $\times$     & Dynamic Spatial Graph & 39.52 \\
    $\checkmark$ & $\checkmark$ & Dynamic Spatiotemporal Graph & 38.54 \\
    \bottomrule
  \end{tabular}%
  
\end{table*}

After further incorporating temporal constraints, the MPJPE decreases significantly from 39.16 mm to 38.54 mm. This indicates that temporal information plays an important regularization role in dynamic relation modeling by constraining the variation of adjacency relationships, thereby improving graph stability over time.

Overall, the results demonstrate that spatial dynamic relation modeling alone is insufficient for stable performance. By jointly modeling spatial and temporal dynamic relations, GCN-DSR enables more stable dynamic relation learning and improves pose estimation accuracy.

\subsubsection{Analysis of RSRM}

To evaluate the effectiveness of the proposed Residual Skeletal Recalibration Module (RSRM), we investigate different recalibration strategies and insertion positions.

Following the original MixTGFormer design, an SE-based channel attention module is used as the baseline recalibration mechanism. Based on this setting, we replace the SE module with RSRM while keeping all other components unchanged for fair comparison. The corresponding results are reported in Table~\ref{tab:rsrm_ablation}.

\begin{table}[htbp]
  \centering
  \caption{Analysis of different recalibration strategies and insertion positions.}
  \label{tab:rsrm_ablation}
  \setlength{\tabcolsep}{3pt} 
  \begin{tabular}{ccc}
    \toprule
    Recalibration Strategy & Insertion Position & MPJPE \\
    \midrule
   SE      & Fusion Head & 39.16 \\
    RSRM    & Fusion Head & 38.85 \\
    RSRM    & Intra-Block & 38.61 \\
    \bottomrule
  \end{tabular}%
  
\end{table}

As shown in Table~\ref{tab:rsrm_ablation}, replacing SE with RSRM under the same insertion position reduces the MPJPE from 39.16 mm to 38.85 mm, demonstrating the effectiveness of the proposed recalibration mechanism. Furthermore, inserting RSRM within each DSR Hybrid Block further improves the performance to 38.61 mm, suggesting that earlier feature recalibration is beneficial for feature interaction and representation learning.

Overall, the results demonstrate that RSRM provides consistent performance improvements under different insertion configurations.

\subsubsection{Analysis of Hierarchical Fusion Mechanisms}

To evaluate the effectiveness of the proposed hierarchical fusion strategy, we investigate different intra-block and cross-stream fusion mechanisms.

\begin{table}[htbp]
  \centering
  \caption{Analysis of hierarchical fusion mechanisms.}
  \label{tab:fusion_analysis}
  \setlength{\tabcolsep}{3pt} 
  \begin{tabular}{ccc}
    \toprule
    Intra-Block Fusion & Cross-Stream Fusion & MPJPE \\
    \midrule
     Weighted Summation & Weighted Summation & 39.16 \\
    Weighted Summation & Dynamic Channel Fusion & 38.44 \\
    Learnable Gating & Dynamic Channel Fusion & 38.00 \\
    \bottomrule
  \end{tabular}%
\end{table}

As shown in Table~\ref{tab:fusion_analysis}, replacing the default cross-stream weighted summation with Dynamic Channel Fusion reduces the MPJPE from 39.16 mm to 38.44 mm, demonstrating the effectiveness of channel-wise adaptive feature integration.

Furthermore, replacing the intra-block weighted summation with the proposed learnable gating mechanism further improves the performance to 38.00 mm. This indicates that adaptive balancing between the MHSA branch and the GCN-DSR branch facilitates more effective feature interaction within each DSR block.

Overall, the results demonstrate that the proposed hierarchical fusion strategy improves feature complementarity across different fusion levels.

\subsubsection{Summary}

Overall, the ablation studies validate the effectiveness of the proposed framework from three aspects: dynamic skeletal relation modeling, structure-aware feature recalibration, and hierarchical feature fusion. The results indicate that these components provide complementary benefits and jointly contribute to improved pose estimation performance.

\subsection{Qualitative Analysis}

\subsubsection{Visualization of Dynamic Skeletal Relations}

\begin{figure}[pos=htbp]
  \centering
  \includegraphics[width=0.9\linewidth, height=0.4\textheight, keepaspectratio]{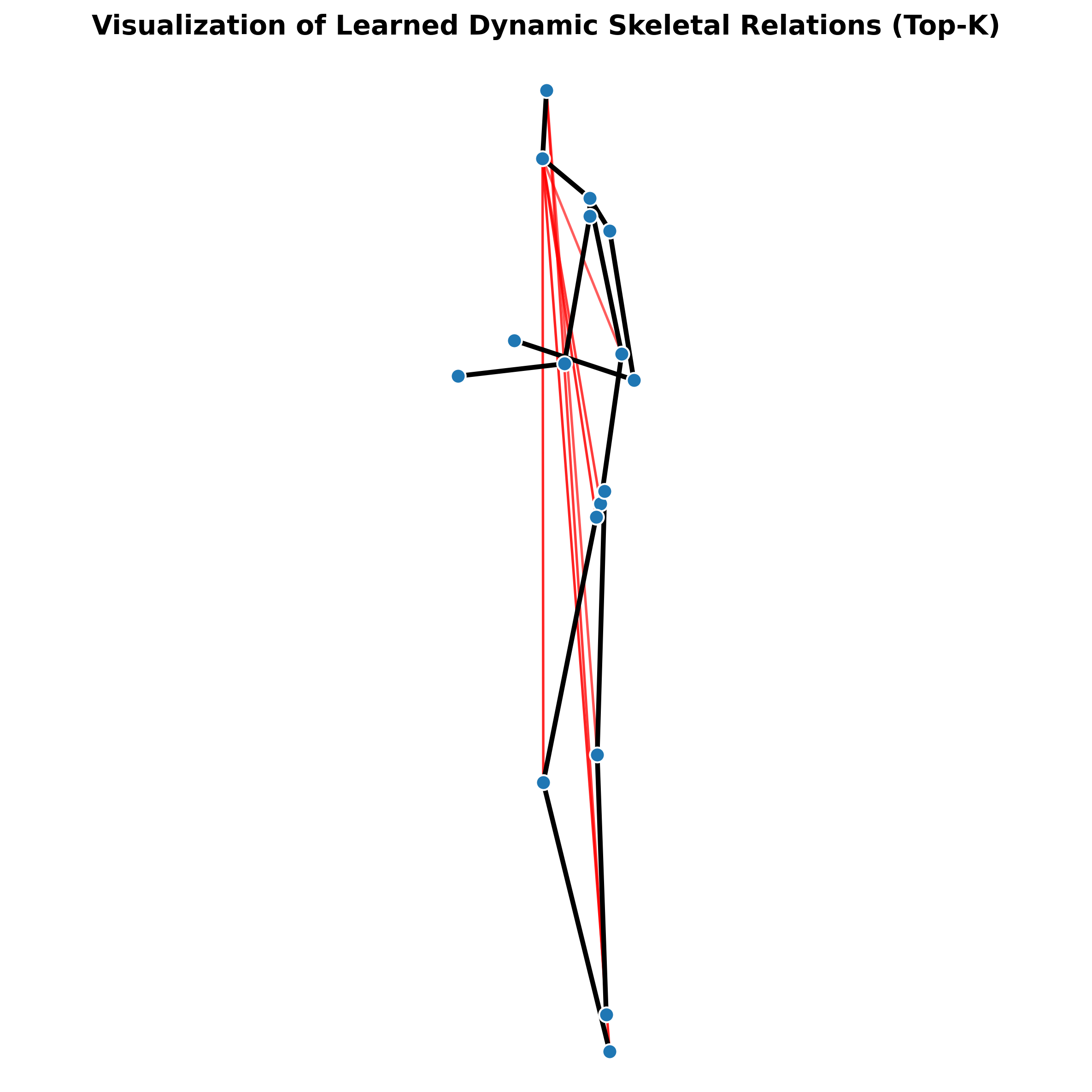}
  \caption{Spatial visualization of learned dynamic skeletal relations (Top-K).}
  \label{fig:skeletal}
\end{figure}

\begin{figure}[pos=htbp]
  \centering
  \includegraphics[width=1\linewidth, height=0.45\textheight, keepaspectratio]{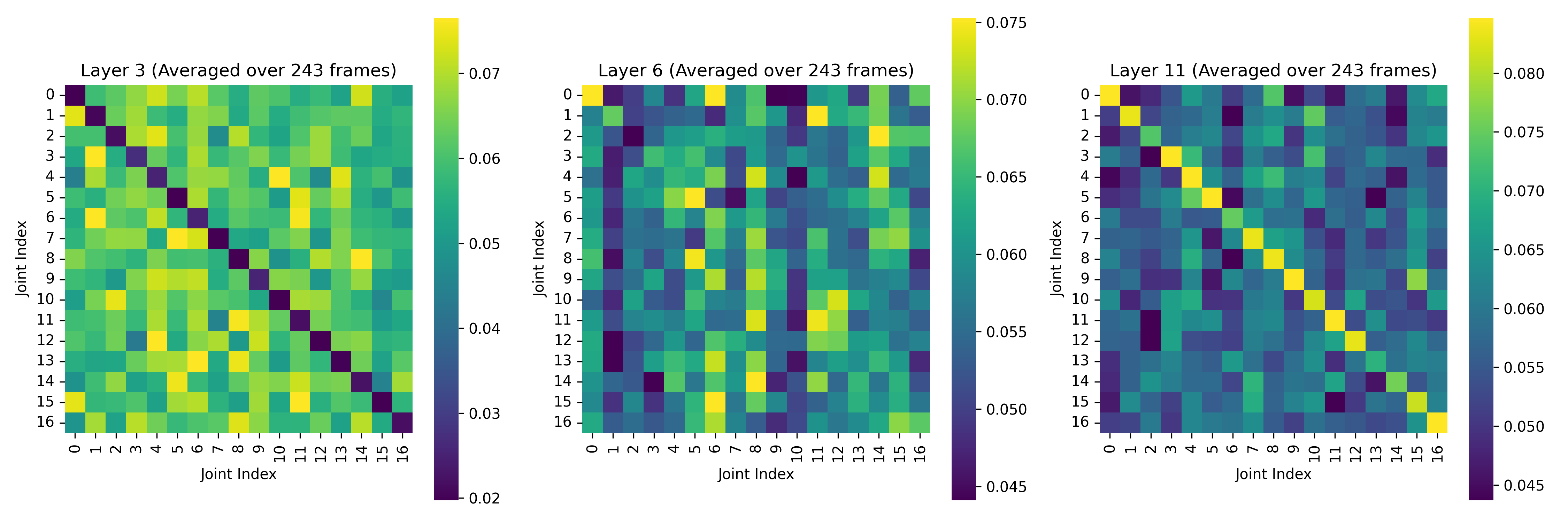}
  \caption{Visualization of the learned skeletal relations at different network layers (Layer 3, Layer 6, and Layer 11).}
  \label{fig:layers}
\end{figure}

To further analyze the effectiveness of the proposed Dynamic Skeletal Relations (DSR), the learned inter-joint relations are visualized from two perspectives, namely spatial visualization and relation matrices, as shown in Fig.~\ref{fig:skeletal} and Fig.~\ref{fig:layers}.

First, from a spatial perspective, as illustrated in Fig.~\ref{fig:skeletal} , the Top-K selected dynamic relations are projected onto the human skeleton for visualization. It can be observed that, compared with the predefined skeletal connections, the learned relations are not limited to local neighborhoods but also capture long-range dependencies across joints. For example, some connections span across the torso and limbs, indicating that the model can adaptively focus on structurally important joints according to the current pose configuration, thereby modeling more global spatial dependencies.

Second, from the perspective of relation matrices, the distribution of dynamic relations at different network layers is analyzed. As shown in Fig.~\ref{fig:layers}, in the shallow layers, the relations are relatively dispersed, reflecting an initial exploration of potential inter-joint dependencies. As the network depth increases, the relation patterns become progressively more concentrated and structured, suggesting that the model gradually strengthens discriminative dependencies between joints.

These results demonstrate that the proposed DSR mechanism is capable of adaptively modeling dynamic skeletal relations with long-range dependencies across different layers, rather than relying on fixed or locally constrained topologies, thereby improving the representation capability for complex human poses.

\begin{figure*}[t]
  \centering
  \includegraphics[width=0.9\linewidth]{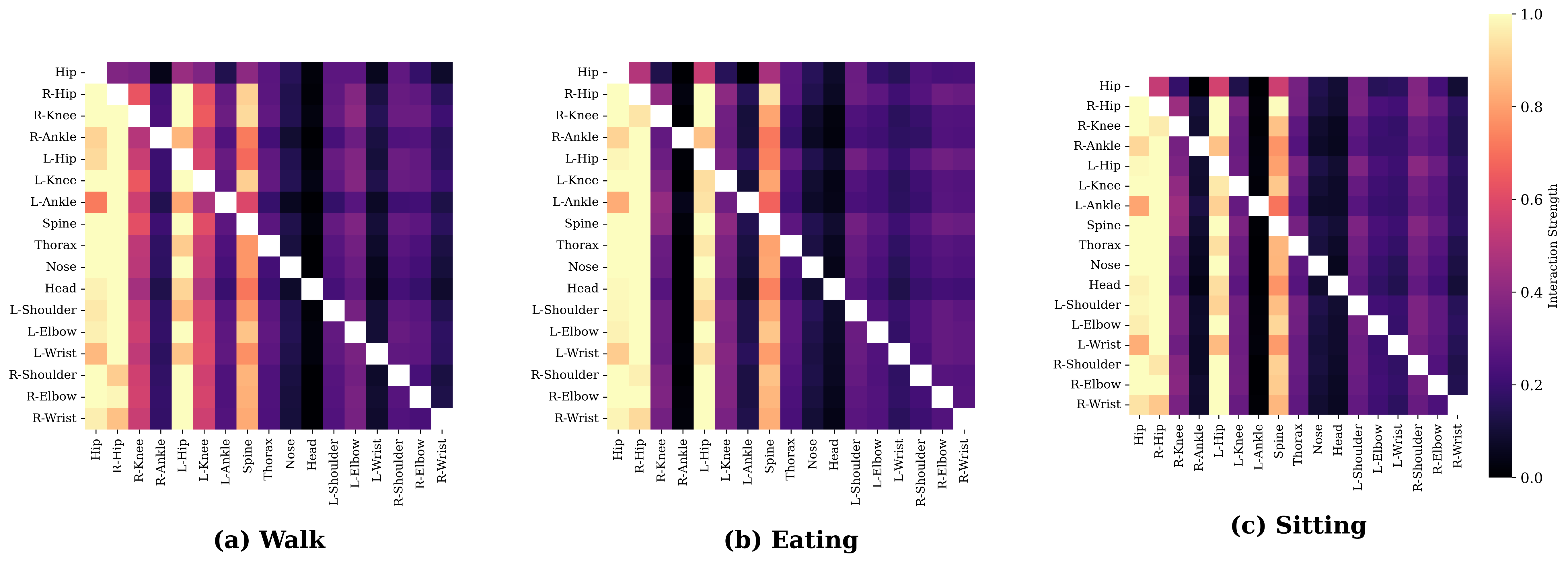}
  \caption{
Visualization of learned dynamic skeletal relations under different actions: (a) Walk, (b) Eating, and (c) Sitting.
}
  \label{fig:action_relations}
\end{figure*}

\subsubsection{Action-wise Dynamic Relation Analysis}

To further investigate whether the proposed DSR mechanism can adaptively capture motion-dependent skeletal dependencies, we visualize the learned relation matrices under different action categories, including walking, eating, and sitting, as shown in Fig.~\ref{fig:action_relations}.

It can be observed that the learned inter-joint relations exhibit distinct distribution patterns across different actions. Specifically, for the walking action, stronger interactions are mainly concentrated among lower-limb joints, reflecting coordinated leg motion during locomotion. In contrast, for the eating action, the learned relations shift toward upper-body joints, especially around the shoulders, elbows, and wrists, which are directly involved in hand-object interaction. For the sitting action, the model emphasizes relatively stable dependencies around the torso and hip regions while suppressing highly dynamic limb interactions.

These observations suggest that DSR adaptively adjusts skeletal relation patterns according to action semantics, enabling motion-dependent structural representation learning.

\subsubsection{Comparison of 3D Pose Estimation Results}

\begin{figure*}[t]
  \centering
  \includegraphics[width=0.5\textwidth, keepaspectratio]{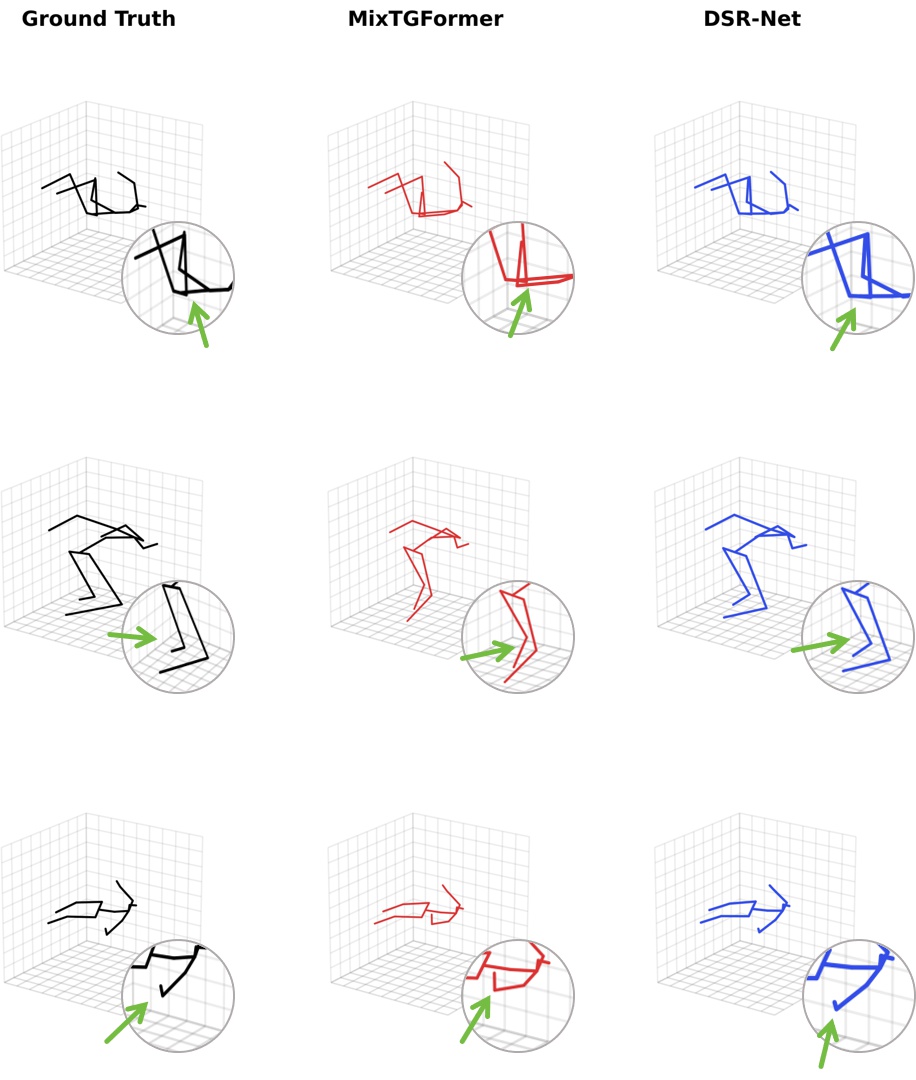}
  \caption{Qualitative comparison of 3D pose estimation results between the proposed method and MixTGFormer on the Human3.6M dataset.}
  \label{fig:skeleton3}
\end{figure*}

To provide a more intuitive evaluation of model performance, Fig.~\ref{fig:skeleton3} presents a qualitative comparison between the proposed method and MixTGFormer on the Human3.6M dataset. It can be observed that in challenging scenarios such as limb crossing, rapid motion, and joint occlusion, MixTGFormer tends to produce joint position deviations or structurally inconsistent skeletons.

In contrast, the proposed method generates more stable and structurally coherent 3D poses. For example, in cases involving arm crossing and large leg movements, the predicted joint connections are clearer, and bone lengths remain consistent, helping reduce structural distortions. These results demonstrate that the proposed dynamic skeletal relation modeling and feature fusion strategy effectively enhance structural modeling capability under complex pose conditions.

\subsubsection{Qualitative Analysis in Challenging In-the-Wild Scenarios}

\begin{figure*}[t]
  \centering
  \includegraphics[width=0.7\textwidth, keepaspectratio]{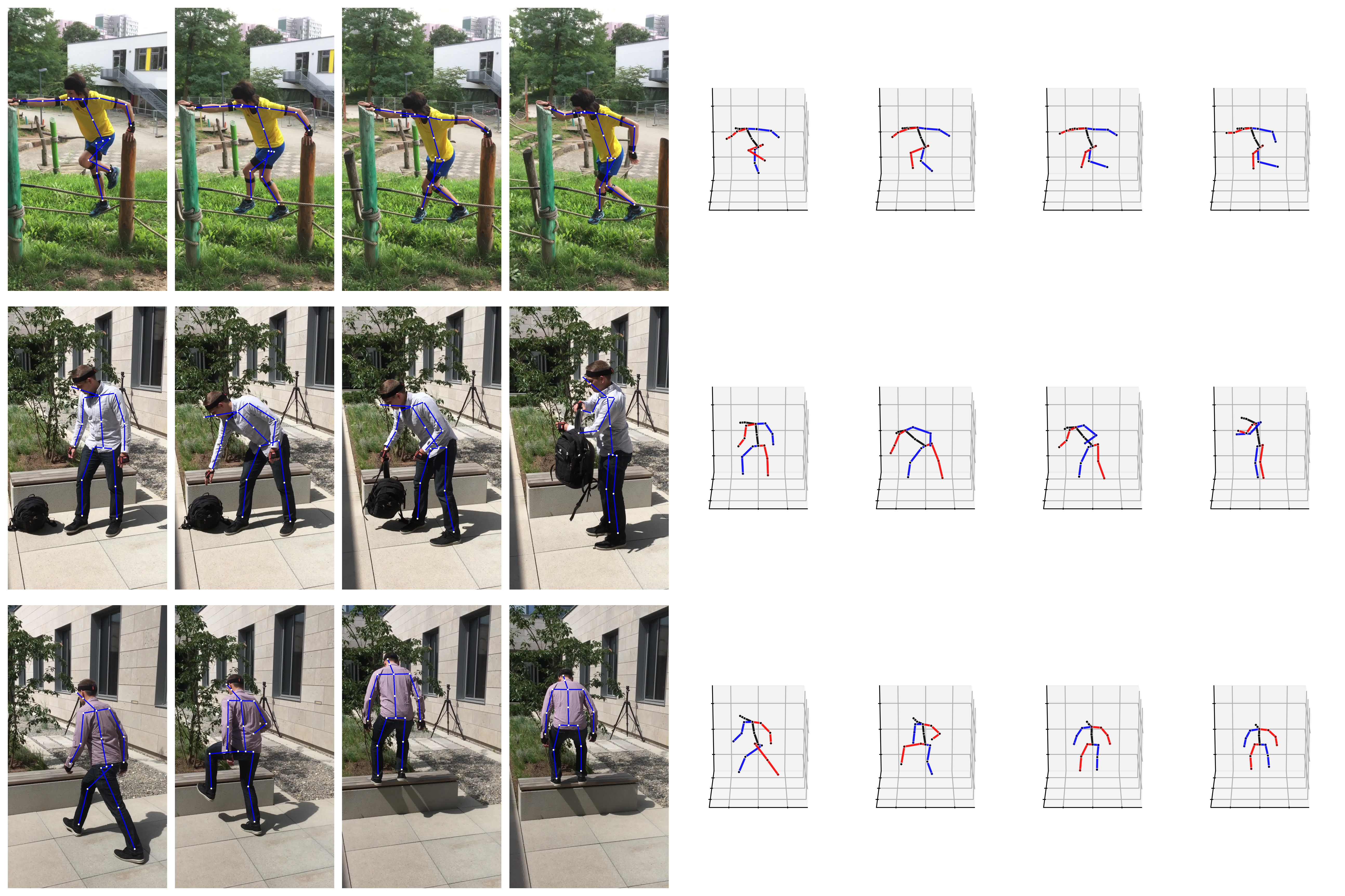}
  \caption{Qualitative results of DSR-Net on the 3DPW dataset under challenging in-the-wild scenarios, including outdoor environments, motion variations, and partial occlusions.}
  \label{fig:vis}
\end{figure*}

Fig.~\ref{fig:vis} presents qualitative results on the 3DPW dataset under challenging in-the-wild conditions. Despite large viewpoint variations, motion diversity, and partial occlusions, DSR-Net preserves anatomically plausible skeletal structures and produces stable pose predictions. These results suggest that the learned dynamic skeletal relations exhibit favorable generalization capability beyond controlled indoor benchmark datasets.

For better visualization, the left and right body parts are highlighted in blue and red, respectively.

\section{Conclusion}

This paper presents DSR-Net, a hybrid Graph-Transformer framework for 3D human pose estimation based on dynamic skeletal relation learning. Unlike conventional methods that rely on fixed skeletal topology, the proposed framework dynamically models motion-dependent inter-joint dependencies, enabling more adaptive structural representation under diverse motion patterns.

To effectively exploit the learned relations, a hierarchical relation-aware fusion strategy is introduced to enhance feature interaction across different modeling branches. In addition, a lightweight Residual Skeletal Recalibration Module (RSRM) is designed to improve structural consistency during feature propagation while maintaining computational efficiency.

Experimental results on Human3.6M and MPI-INF-3DHP demonstrate the effectiveness of dynamic skeletal relation modeling and hierarchical feature fusion for 3D human pose estimation. The proposed framework achieves competitive performance while providing a flexible mechanism for adaptive structural representation learning.

Future work will focus on developing more efficient dynamic relation learning mechanisms and exploring long-range motion modeling under lightweight architectures. We also plan to investigate the applicability of the proposed framework to more challenging real-world human motion analysis scenarios.

\section*{CRediT authorship contribution statement}
\textbf{Fan Li:} Conceptualization, Methodology, Software, 
Formal analysis, Investigation, Writing -- original draft, 
Visualization.
\textbf{Jian Xiang:} Supervision, Resources, Project 
administration, Writing -- review \& editing, 
Funding acquisition.
\textbf{Linlin Xue:} Supervision, Writing -- review \& 
editing, Funding acquisition.
\textbf{Yuwei Gao:} Validation, Data curation, Investigation.

\section*{Declaration of competing interests}

The authors declare that they have no known competing financial interests or personal relationships that could have appeared to influence the work reported in this paper.

\section*{Data availability}

Data will be made available on request.

\section*{Acknowledgments}

This work was supported by the Leading Goose R\&D Program of Zhejiang Province (Grant No. 2025C02045).

\bibliographystyle{cas-model2-names}
\bibliography{ref}
\end{document}